\documentclass[conference]{IEEEtran}
\IEEEoverridecommandlockouts
\usepackage{cite}
\usepackage{amsmath,amssymb,amsfonts}
\usepackage{algorithmic}
\usepackage{graphicx}
\usepackage{textcomp}
\usepackage[table]{xcolor}
\usepackage{xcolor}
\def\BibTeX{{\rm B\kern-.05em{\sc i\kern-.025em b}\kern-.08em
    T\kern-.1667em\lower.7ex\hbox{E}\kern-.125emX}}

\usepackage{tabularx}
\usepackage{url}
\usepackage{caption}

\usepackage{dblfloatfix}

\usepackage{soul}
\usepackage{courier}
\usepackage[T1]{fontenc}
\usepackage{epstopdf}
\newcolumntype{Y}{>{\centering\arraybackslash}X}
\usepackage{collcell}
\usepackage{xfp}

\usepackage{listings}                       

\makeatletter
\newcommand\fs@spaceruled{\def\@fs@cfont{\bfseries}\let\@fs@capt\floatc@ruled%
  \def\@fs@pre{\vspace{0.5\baselineskip}\hrule height.8pt depth0pt \kern2pt}%
  \def\@fs@post{\kern2pt\hrule\relax}%
  \def\@fs@mid{\kern2pt\hrule\kern2pt}%
  \let\@fs@iftopcapt\iftrue}%
\makeatother
\begin{document}

\title{What and How Are We Reporting in HRI?\\ A Review and Recommendations for Reporting Recruitment, Compensation, and Gender \\
\thanks{*Equal Contribution}
}

\author{
  \IEEEauthorblockN{Julia R. Cordero*}
  \IEEEauthorblockA{\textit{Viterbi School of Engineering} \\
    \textit{University of Southern California}\\
    Los Angeles, USA \\
    jrcorder@usc.edu}
  \and
  \IEEEauthorblockN{Thomas R. Groechel*}
  \IEEEauthorblockA{\textit{Viterbi School of Engineering} \\
    \textit{University of Southern California}\\
    Los Angeles, USA \\
    groechel@usc.edu}
  \and
 \IEEEauthorblockN{ Maja J. Matari\'c}
  \IEEEauthorblockA{\textit{Viterbi School of Engineering} \\
    \textit{University of Southern California}\\
    Los Angeles, USA \\
    mataric@usc.edu}
}

\IEEEaftertitletext{\vspace{-3\baselineskip}} 

\maketitle

\begin{abstract}
Study reproducibility and generalizability of results to broadly inclusive populations is crucial in any research.  Previous meta-analyses in HRI have focused on the consistency of reported information from papers in various categories.  However, members of the HRI community have noted that much of the information needed for  reproducible and generalizable studies is not found in published papers. We address this issue by surveying the reported study metadata over the past three years (2019 through 2021) of the main proceedings of the International Conference on Human-Robot Interaction (HRI) as well as alt.HRI. Based on the analysis results, we propose a set of recommendations for the HRI community that follow the longer-standing reporting guidelines from human-computer interaction (HCI), psychology, and other fields most related to HRI. Finally, we examine three key areas for user study reproducibility: recruitment details, participant compensation, and participant gender. We find a lack of reporting within each of these study metadata categories: of the 236 studies, 139 studies failed to report recruitment method, 118 studies failed to report compensation, and 62 studies failed to report gender data.  This analysis therefore provides guidance about specific types of needed reporting improvements for HRI. 
\end{abstract}


\section{Introduction}
\label{sec:introduction}
\textit{Reproducibility}--the ability to duplicate a prior study's procedure and obtain the same results\cite{cacioppo2015social}-- is critical for human-subject studies that aim to be \textit{generalizable}, i.e., applied to different populations and settings\cite{carminati2018generalizability}. Human-robot interaction (HRI), similar to its related fields human-computer interaction (HCI) and psychology, conducts studies in critical areas such as health and education, increasing the importance of reproducibility and generalizability of its results. Reproducibility has been a relatively prevalent topic for HRI; the International Conference on Human-Robot Interaction itself introduced a ``Reproducibility in Human-Robot Interaction'' track in 2020, focused on reproducing prior HRI work \cite{hri2020mainproceedings}. HRI researchers have conducted studies to replicate findings from previous work \cite{strait2020three, ullman2021challenges}. However, despite these efforts, a lack of reproducibility for studies has been recognized as a “crisis” in psychology and the sciences in general \cite{baker2015first, baker2016reproducibility}. 

HRI falls under this call to action to bridge the gap in research reproducibility. However, in order for a study to be reproducible and generalizable, it must report metadata (e.g., participant demographics) comprehensively and in enough detail for other researchers to evaluate it in context and aim to reproduce the reported results. Hence, HRI must aim to meet a higher standard of rigor in study reporting.

To determine best practices for reporting studies with participants, we took inspiration from the method of review presented in ``Four Years in Review: Statistical Practices of Likert Scales in Human-Robot Interaction Studies'' \cite{schrum2020four}. Accordingly, we first conducted a review of the reporting standards in related fields (HCI, psychology). Then, to evaluate the current level of study reporting in HRI, we collected data on and examined metadata being reported and not reported. We found that two metadata categories, participant recruitment and participant compensation, were reported infrequently or with an inadequate amount of contextual information. We also found participant gender, while reported in $\approx73.7$\% of studies, was reported in a way that did not account for the full complexity of gender considerations. In HRI, having a contextualized understanding of what participants experienced is critical for being able to reproduce a study and then generalize results to a broader, inclusive population.

The following are the main contributions of this work.
First, we contribute a review of the existing literature for reporting guidelines and standards in psychology, HCI, and general qualitative research. While some of those guidelines apply to specific types of studies (e.g., randomized control trials in medicine), all of the surveyed literature agreed on minimum details to be reported. We provide recommendations that apply best to the user studies that are prevalent in the HRI community.

Second, we contribute a review of papers published in the International Conference on Human-Robot Interaction from 2019 through 2021 with regard to reported metadata (both main proceedings and alt.HRI). We reviewed 174 papers, 155 of which contained at least one study, totalling 236 studies.  We found that 139 ($\approx$58.9\%) studies did not report details about participant recruitment, 118 ($\approx$50.0\%)  studies did not report any details about compensation, and 62 ($\approx$26.3\%) studies did not report details about participant gender. Based on these findings, we argue that there is a need for reporting recommendations for the HRI community.

\vspace{2mm}
\noindent\fbox{%
    \parbox{0.97\columnwidth}{%
        \textbf{Nota Bene}: Similar to ``Four Years in Review: Statistical Practices of Likert Scales in Human-Robot Interaction Studies'' \cite{schrum2020four}, we admit that we have not employed best practices in our own prior work. Our goal for this paper is not to disparage the field, but instead to highlight the lack of reporting of study metadata. We hope to improve the rigor of our own and others’ work so that we can all contribute more confidently to the reproducibility and generalizability of HRI studies.
    }%
} 

\section{Literature Review \& Best Practices}
\label{sec:reviewandrec}
HRI studies by definition involve human behavior. Each study participant's behavior is necessarily  influenced by the participant's identity and experiences with the researcher and the experiment methodology and context. Hence, explicitly reporting what participants experience allows for a more complete interpretation of results and study reproducibility \cite{birhane2021impossibility}. Reporting guidelines have been proposed and adopted in other fields that make use of qualitative user studies; while some guidelines apply best to the practices in particular fields (such as a randomized control trial in medicine), HRI shares a dependency on human subjects and thus can benefit from the examination of minimum information reporting already established in longer-standing related fields.

\subsection{Recruitment}
\textbf{Recruitment Method} Previous work has highlighted the importance of the way that researchers reach potential participants. The textbook \textit{Qualitative HCI Research} states that the recruitment strategy may affect “quality, reliability, or generalizability” of a study and thus should be considered important to report. This acknowledges the role recruitment methods play in a study’s methodology and, by extension, reproducibility \cite{blandford2016qualitative}. 

Recruitment reporting is covered in a number of reporting guidelines; for example, Standards for Reporting Qualitative Research (SRQR) and Consolidated Criteria for Reporting Qualitative Research (COREQ), both written for qualitative health research~\cite{altman2008equator}, state that the manner of selecting and approaching participants needs to be explicitly reported \cite{o2014standards,tong2007consolidated}. Describing the recruitment method is also a part of the reporting checklist proposed for psychology studies by the American Psychological Association (APA)~\cite{appelbaum2018journal} and Consolidated Standards of Reporting Trials (CONSORT)~\cite{bennett2005consolidated, altman2008equator}.

Reporting the recruitment method has larger implications beyond simply detailing the flow of participants through the study; previous work suggests that the recruitment method should also include details about compensation~\cite{pater2021standardizing} and participant gender~\cite{offenwanger2021diagnosing}.

\textbf{Recruitment Setting}
Not all guidelines provide recommendations about reporting recruitment settings (e.g., \cite{o2014standards, tong2007consolidated}), CONSORT, the guidelines written for randomized control trials (RCTs) suggests a standard for reporting recruitment settings \cite{bennett2005consolidated}. The American Psychological Association (APA) guidelines also include reporting the recruiting context (e.g., location, time period) in qualitative research \cite{appelbaum2018journal}. We believe that reporting the setting in which participants are recruited provides important context about data sources for future researchers aiming to reproduce a study.

\textbf{Inclusion and Exclusion Criteria}
Inclusion criteria are defined as the broad definition of eligibility for a study, while exclusion criteria are defined as any criteria that remove any potentially eligible participants from the pool; both are intended to be applied before the study takes place \cite{wright2006importance}. Although this distinction is drawn clearly and recommended to be reported as two separate items by some works (\cite{tong2007consolidated, o2014standards, toerien2009review, wright2006importance}), CONSORT states only that “eligibility criteria” should be reported \cite{bennett2005consolidated}. When reporting decisions made for inclusion and exclusion criteria, SRQR recommends stating the justification for the decision in addition to stating the criteria themselves~\cite{o2014standards}.

\textbf{Ethical Approval}
SRQR recommends reporting ethical approval \cite{o2014standards}, while COREQ and CONSORT do not provide recommendations on that point \cite{bennett2005consolidated, tong2007consolidated}. More broadly, Qualitative HCI recommends reporting ethical considerations and how they did or did not affect decisions in recruitment and study design \cite{blandford2016qualitative}. This perspective also agrees with SRQR’s recommendation to provide justification for study decisions, as well as to report the decision made~\cite{o2014standards}.

\textbf{Recommendation:} 
All of the guidelines we reviewed stated which details should be included in a study report; SRQR made the additional recommendation that explanations accompany the reported study metadata so as to provide visibility into the researcher’s role and the study’s aims. While this was not a consensus among guidelines included in our literature review, we believe that the additional information further enhances study reproducibility.

Based on the relevant literature review, we recommend: 
\begin{itemize}
\item Reporting ethical approval from the relevant regional study approval board;
\item Explaining what ethical considerations were factored into study design;
\item Reporting the recruitment method and setting and explaining why the chosen recruitment strategy fits the aims of the study. (For an HRI conference paper example of reporting the strategy with an explanation, see ``The Effects of a Robot's Performance on Human Teachers for Learning from Demonstration Tasks'' \cite{hedlund2021effects}.)
\item Reporting both the inclusion criteria (the population for study) and exclusion criteria (what causes a person from the population to be disqualified from the study) and explaining the justification for choosing them for the study.
\end{itemize}

\subsection{Compensation}
While there is little specific guidance from various reporting guidelines on whether and how to report study participant compensation, there are broader recommendations to provide explanations of ethical considerations in study design \cite{blandford2016qualitative}. In the healthcare research community, participant compensation is considered an ethical issue~\cite{brown2018need, lynch2019association, ripley2006review}.  Additionally, previous work in HCI has called for more standardized reporting of participant compensation in order to increase study replicability~\cite{pater2021standardizing}.

\textbf{Form and Amount of Compensation} Some reporting standards do not provide recommendations on reporting participant payment (e.g., \cite{o2014standards, bennett2005consolidated, tong2007consolidated}). However, ethics boards (e.g., Institutional Review Boards) typically require that researchers report participant incentives, which casts compensation as an ethical factor to be reported~\cite{compensationoregon}. The APA recommends that researchers “describe any incentives or compensation” in their study reports \cite{appelbaum2018journal}; in HCI, a proposed standard is to report both the form and amount of participant compensation~\cite{o2014standards}. In the context of clinical trials for HIV research, there has been a call to record compensation for studies for the benefit of other researchers to maintain consistency when replicating the same study~\cite{brown2018need}.

\textbf{Location and Duration}
The proposed HCI standard \cite{pater2021standardizing} is to report location and duration along with compensation form and amount so as to contextualize the value of the incentives. The discussion of tracking participant compensation also references the role of location in contextualizing the amount of payment \cite{brown2018need}. No explicit recommendations are provided in CONSORT on reporting study location or duration \cite{bennett2005consolidated}; COREQ, however, suggests a standard of reporting the duration of time participants are studied \cite{tong2007consolidated}. SRQR guidelines recommend reporting context (defined as “setting/site and salient contextual factors”) and the rationale for the context \cite{o2014standards}; this recommendation agrees with APA’s emphasis on the role of context in human subjects studies and the need to report the study context~\cite{appelbaum2018journal}. 

\textbf{Recommendation:} Reporting compensation goes beyond reporting whether participants were paid; it is important to include amounts in order to understand the full impact of the incentive. Reporting the time that participants spent in the study and where they were located is just as important as part of the context. Although there is a lack of consensus of the importance of reporting participant compensation, we believe that the standards proposed in HCI can benefit HRI most due to the similarity in study structure.

Based on the relevant literature review, we recommend:
\begin{itemize}
\item Reporting the form (e.g., voucher, course credit, gift card) and value of the compensation;
\item Reporting the location of the study to further contextualize the value of compensation;
\item Reporting the full study duration including data collection phases (e.g., ``participants performed tasks with the robot for 20 minutes'') and total time spent in the study (e.g., ``the participant was at the research facility for 1.5 hours'').
\end{itemize}

\subsection{Gender}
Unlike reporting recruitment and compensation, reporting participant gender relies on self report. Therefore, the ongoing discussion about reporting gender considers ways that researchers use to ask participants to report their gender. Below, we review common challenges in collecting and reporting participant gender information, and a set of different researcher perspectives on how to address those challenges.

\textbf{When is reporting participant gender harmful?} Previous work has addressed the importance of reporting patricipant gender to study reproducibility~\cite{baker2015first, baker2016reproducibility}. However, such reporting can pose some risks for participants. During demographics surveys, for example, forcing a participant to answer a question about gender can cause distress if they are unable to give a definite answer~\cite{scheuerman2020hci, spiel2019better}. Reporting gender data transparently can also carry risk for some under-represented participants. The HCI Gender Guidelines describe a scenario in which a transgender participant can have their identity revealed if not enough of their demographic data are removed, potentially placing them in a dangerous situation if recognized~\cite{scheuerman2020hci}. 

\textbf{How do we avoid assuming that gender is binary?} HRI researchers have called for collecting gender data in a way that is nuanced and inclusive, as is done in existing gender reporting guidelines in HCI \cite{offenwanger2021diagnosing}. The HCI Gender Guidelines state to “include transgender and non-binary into \ldots definitions of gender” \cite{scheuerman2020hci}. They also suggest leaving participant gender to be disclosed in an open-ended answer box in a survey and to allow participants to report multiple genders. Specifically, the HCI guidelines recommend allowing aggregate reports of gender to exceed 100\% in order to include multiple gender categories per participant~\cite{scheuerman2020hci}. This recommendation is further supported by Hughes~\cite{hughes2016rethinking} stating that an “open-ended response item helps with [reporting multiple genders] because participants with multiple gender identities can write multiple answers”.

\textbf{What is the role of the "other" gender option?} The designation ``other'' has been used in demographics surveys as an alternative to gender binary options (i.e., man, woman). Hughes~\cite{hughes2016rethinking} suggests using “other” as a catch-all option for participants who are not otherwise represented in a survey. However, others (e.g., \cite{scheuerman2020hci, offenwanger2021diagnosing, spiel2019better}) recommend against using the term in surveys and reports because “other” can imply existing outside of the accepted norm and fails to capture the nuance and complexity of gender~\cite{offenwanger2021diagnosing}.

\textbf{Should a free-response option replace "other"?} Offering a free-response text box to specify gender, without additional options to select, is acknowledged as a desirable option for being as inclusive as possible  \cite{hughes2016rethinking, scheuerman2020hci, spiel2019better}. However, one cited limitation of this approach is that a larger participant pool will make coding all responses infeasible for researchers. While using the free-response option may still be preferable for smaller samples \cite{scheuerman2020hci}, a suggested alternative for large pools is to provide five options (woman, man, non-binary, prefer not to disclose, and prefer to specify) and an accompanying free-response box \cite{spiel2019better}. This solution is imperfect, and may not be necessary as modern spreadsheets can sort and count the number of unique instances (e.g., ``female'' = 10 or ``non-binary'' = 15) from free text with minimal technological expertise.

\textbf{Should participants be reported as females/males, or women/men?} The HCI Gender Guidelines acknowledge the complexity of this consideration; gender (woman, man) and sex (female, male) are considered by some to be separate definitions, while others consider differentiating between gender and sex to exclude the identities of trans people. These guidelines suggest avoiding using “male” and “female” (unless reporting a medical study)~\cite{scheuerman2020hci}. A similar recommendation is to use “woman” and “man” as options on a gender survey as opposed to “female” and “male”~\cite{spiel2019better}. Other recommendations have assumed a differentiation between gender and sex but also ultimately recommend using ``woman'', ``man'', ``cisgender'', and ``transgender''~\cite{hughes2016rethinking}.

\textbf{Recommendation:}
Discussing gender is a complex and difficult task in any context; it is even more so in a human subjects study in which researchers must respect participant confidentiality. While no set of guidelines for describing gender will be complete, and each recommendation will need to evolve as new perspectives are raised, we believe that the following recommendations best capture a level of reporting that facilitates generalizability and reproducibility.

Based on the relevant literature review, we recommend:
\begin{itemize}
    \item Avoiding the binary gender assumption. If reporting gender in aggregate, allow the total percentage of gender categories to exceed 100\% to accommodate participants of multiple genders. Explicitly report each category of gender as opposed to assuming binary gender as the default (e.g., ``N=10; 5 participants were female.'');
    \item Avoiding requiring a response to a gender survey to mitigate potential participant anxiety if they are questioning their gender;
    \item Anonymizing gender data to mitigate the chance of revealing a participant’s identity;
    \item Using an optional free-response text box where participants can write in their genders. To help with coding for large pools, use modern spreadsheet or programming tools to automatically sort and count free response text;
    \item Avoiding the term ``other'' as an option in gender surveys and as a category when reporting gender;
    \item Avoiding  ``female/male'' as an option in gender surveys and as descriptors when reporting gender.
\end{itemize}

We acknowledge that following these recommendations can occupy a significant amount of space in space-limited papers. Researchers can opt to include a more through report in supplemental materials; for an HRI conference paper example see ``Assessing and Addressing Ethical Risk from Anthropomorphism and Deception in Socially Assistive Robots''~\cite{winkle2021assessing} for how to include such materials or see ``What Should Robots Feel Like?''~\cite{mcginn2020should} for an example of how to report such information in a compact table. 

Given the above recommendations, we analyzed papers from the International Conference on Human-Robot Interaction (HRI) for the past three years (2019, 2020, 2021), coding for study metadata in an attempt to quantify what is, and what is not, being reported.  This analysis aims to highlight the areas of specific improvements in HRI data reporting.

\section{Review of HRI Papers}
We reviewed and coded the papers from the 2019, 2020, and 2021 International Conference on Human-Robot Interaction main proceedings as well as alt.HRI (a paper track aimed at ``pushing the boundaries of HRI'', published in the companion proceedings for 2020~\cite{althri20} and 2021~\cite{althri21}). We reviewed 174 papers, 155 of which contained at least one study, totalling 236 studies.  We chose to review the last three years (2019 through 2021) as a large and representative enough sample size of papers for analysis. One human coder performed all data annotation for this review.

\subsection{Coding Methodology and Criteria}
We collected data by reading through each paper. For each paper, we performed a combined manual and automated key term search for different columns, as described below (e.g., ``rewarded'') and marked the columns based on the reported data found in the paper. Next, we read the paper fully to code any categories missed by the first search pass, and to check that all data recorded by the first search were correct.

We determined how many, if any, studies the paper contained. A study qualified if the paper reported participants, a study procedure was described, data were collected from those participants, and the data were used or reported. Pilot studies were included if they fit these criteria. We performed an analysis for three categories of study participant metadata: recruitment, compensation, and gender.

The recruitment method was coded and categorized if papers explicitly stated \textbf{how participants were recruited} (e.g., flyers, social media, an outside company). If a paper stated ``we recruited university students'' but omitted how they were recruited, it was coded as not having reported the recruitment method. For online studies conducted via an external online platform (e.g., Amazon Mechanical Turk, Prolific), the recruitment method was implied and thus coded as reported. Online studies run by a university directly, however, had to state how their participants were recruited (e.g., emailed survey to engineering email list) to be coded as reported. 

We also coded the population studied as convenience and non-convenience sampling. We define convenience sampling as any population strictly from a college or university, or recruited by an external online platform (e.g., Amazon Mechical Turk, Prolific). Convenience sampling was coded for anything explicitly stated as \textbf{``convenience sampling'', college/university students (e.g., ``university students''), or implied by the recruitment tool (e.g., ``Amazon Mechanical Turk'')}. Non-convenience sampling was coded when it was either explicitly stated or implied (e.g., ``we worked with grade-school children, hired clinical experts''). Studies were coded as a ``mixture'' if they contained both university and non-university students (e.g., reported as recruiting from the university and surrounding area). Not reported was coded for studies that did not fall into any of the above categories and thus no supported assumptions could be made.


Acknowledgement of ethics board approval was also coded for when a paper explicitly stated \textbf{any acknowledgement of ethics board approval}. While encouraged, specific ethics board approval identifiers (e.g., ``all study materials were reviewed and approved by a University ethics board under application UP-123456'') were not required; a statement regarding the approval of the study was considered sufficient.

Compensation was coded and categorized when papers explicitly stated \textbf{any direct benefit to the participant or a statement of no direct benefit}. Automated search terms included ``reward'', ``award'', ``receive'', ``given'', ``compensated'', and ``paid''. The study procedure sections were also carefully read to check for other wordings of compensation or statements about the participants not being paid. If none of those were found, compensation was coded as not reported. 

Gender was coded and categorized when papers explicitly stated \textbf{any participant gender information}. This included studies that reported only a single gender (e.g., ``N=10; 5 participants were male.''). The number for each distinct gender reported was recorded. In the case of single gender reporting, no other gender was recorded (i.e., no assumptions were made for binary gender). The categories of reported gender information were ``female'', ``non-binary``, ``other'', ``diverse gender'', ``transgender'', ``male'', and ``unidentified''. 


Online studies were also coded and categorized when papers explicitly stated they used an online platform (e.g., Amazon Mechanical Turk, ``online study''). We analyzed sets of data. The first set contained all studies. The second set examined the data when online studies were removed. There were a total of 236 studies of which 73 were online studies ($\approx$31.0\%) and 163 were non-online ($\approx$69.0\%). The three HRI conferences had the following numbers of online and non-online studies: HRI'19 (11,50); HRI'20 (23,80); and HRI'21 (39,34).

\subsection{Recruitment, Compensation, and Gender Frequency}
\begin{figure}[h] 
  \centering
  \includegraphics[width=\columnwidth]{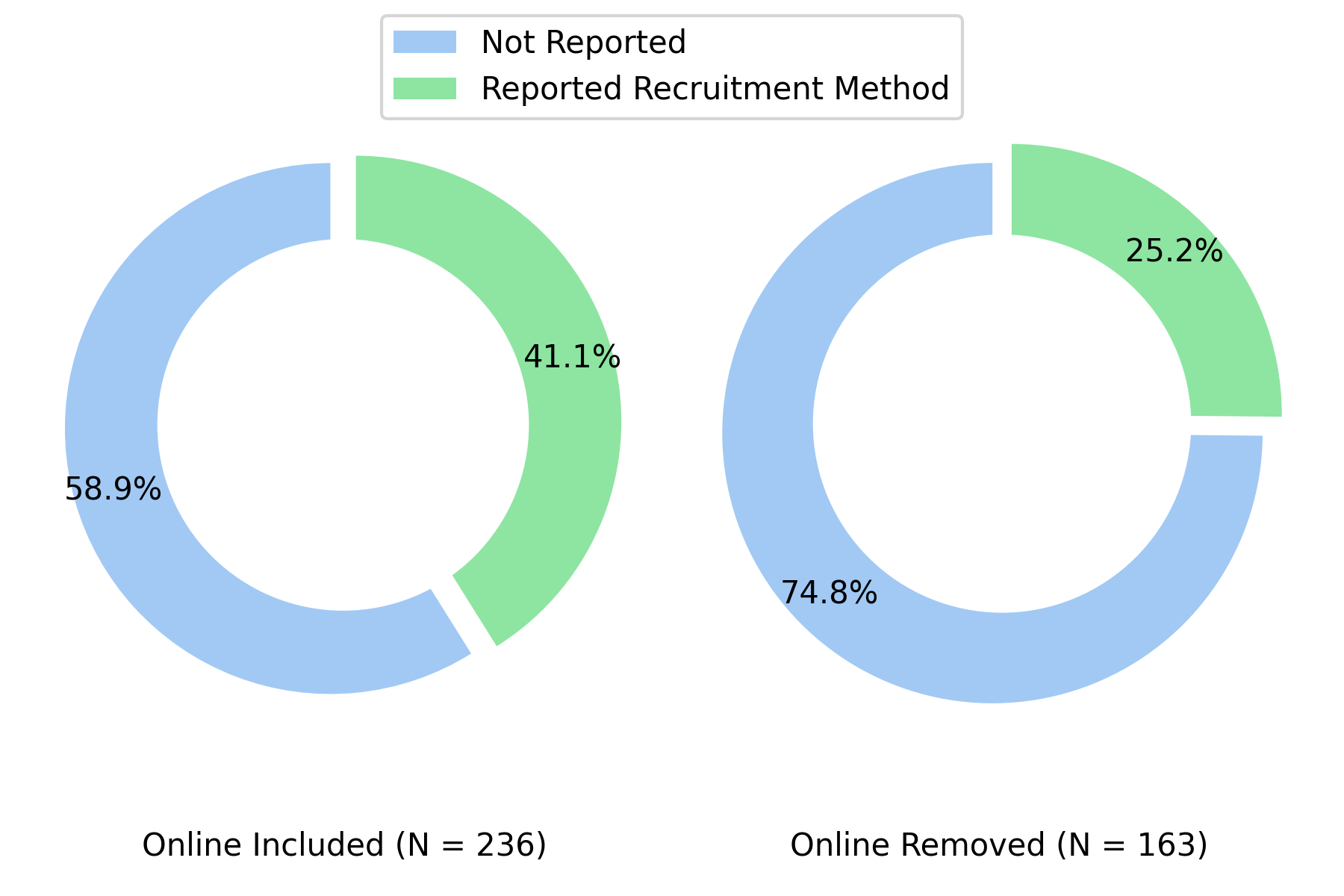}
  \caption{Papers that explicitly stated how participants were recruited (e.g., flyers, social media, an outside company) and papers that did not.} 
  \label{fig:recruimentmethod} 
\end{figure}

\begin{figure}[h] 
  \centering
  \includegraphics[width=\columnwidth]{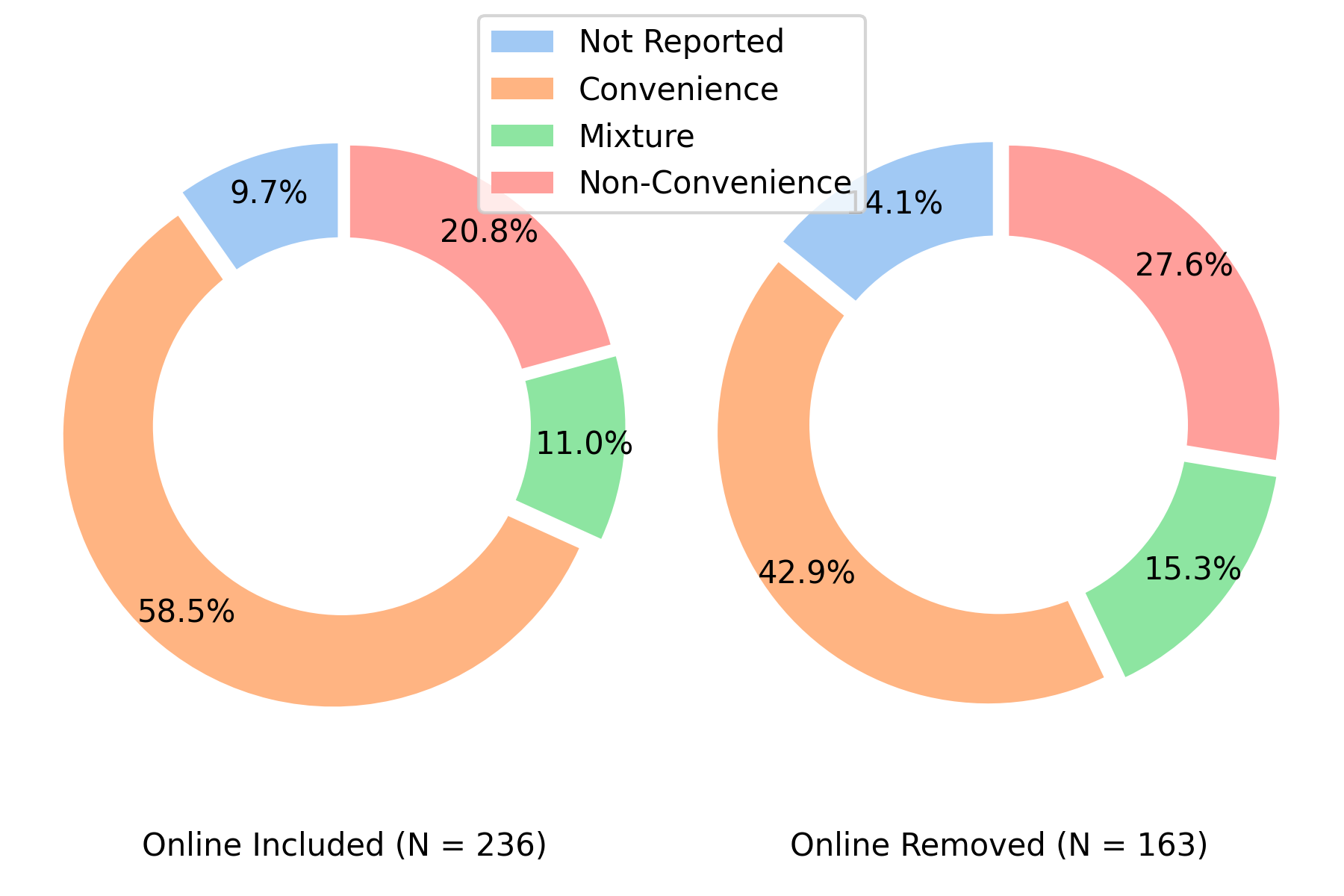}
  \caption{Papers that reported the participant population (convenience, mixture, or non-convenience), either explicitly or implied via the recruitment method (e.g., ``we recruited university students'') and those that did no participant population reporting.} 
  \label{fig:convenience} 
\end{figure}

\begin{figure}[h] 
  \centering
  \includegraphics[width=\columnwidth]{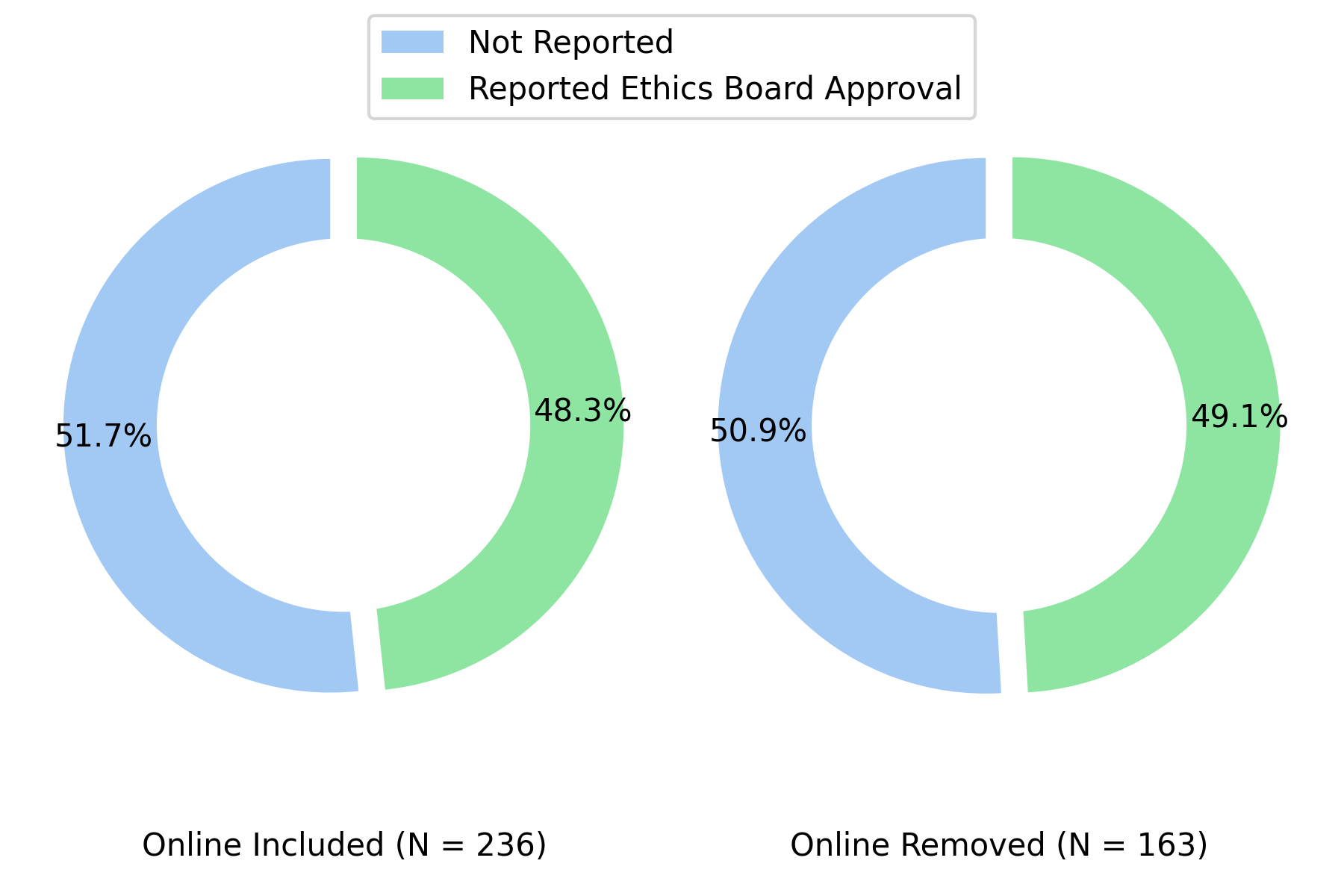}
  \caption{Papers that explicitly stated an acknowledgement of ethics board approval and papers that did not.} 
  \label{fig:ethicsboard} 
\end{figure}

 A total of 97 studies reported recruitment method. When online studies were removed, 41 studies reported recruitment method. For population studied, 138 were coded as convenience, 26 were coded as a mixture, 49 were coded as non-convenience, and 23 did not report their population. When online studies were removed, 70 were coded as convenience, 25 were coded as a mixture, 45 were coded as non-convenience, and 23 did not report their population. A total of 114 studies reported ethics board approval. When online studies were removed, 80 studies reported ethics board approval. Recruitment reporting summaries can be found in Fig. \ref{fig:recruimentmethod}-\ref{fig:ethicsboard}.

\begin{figure}[h] 
  \centering
  \includegraphics[width=\columnwidth]{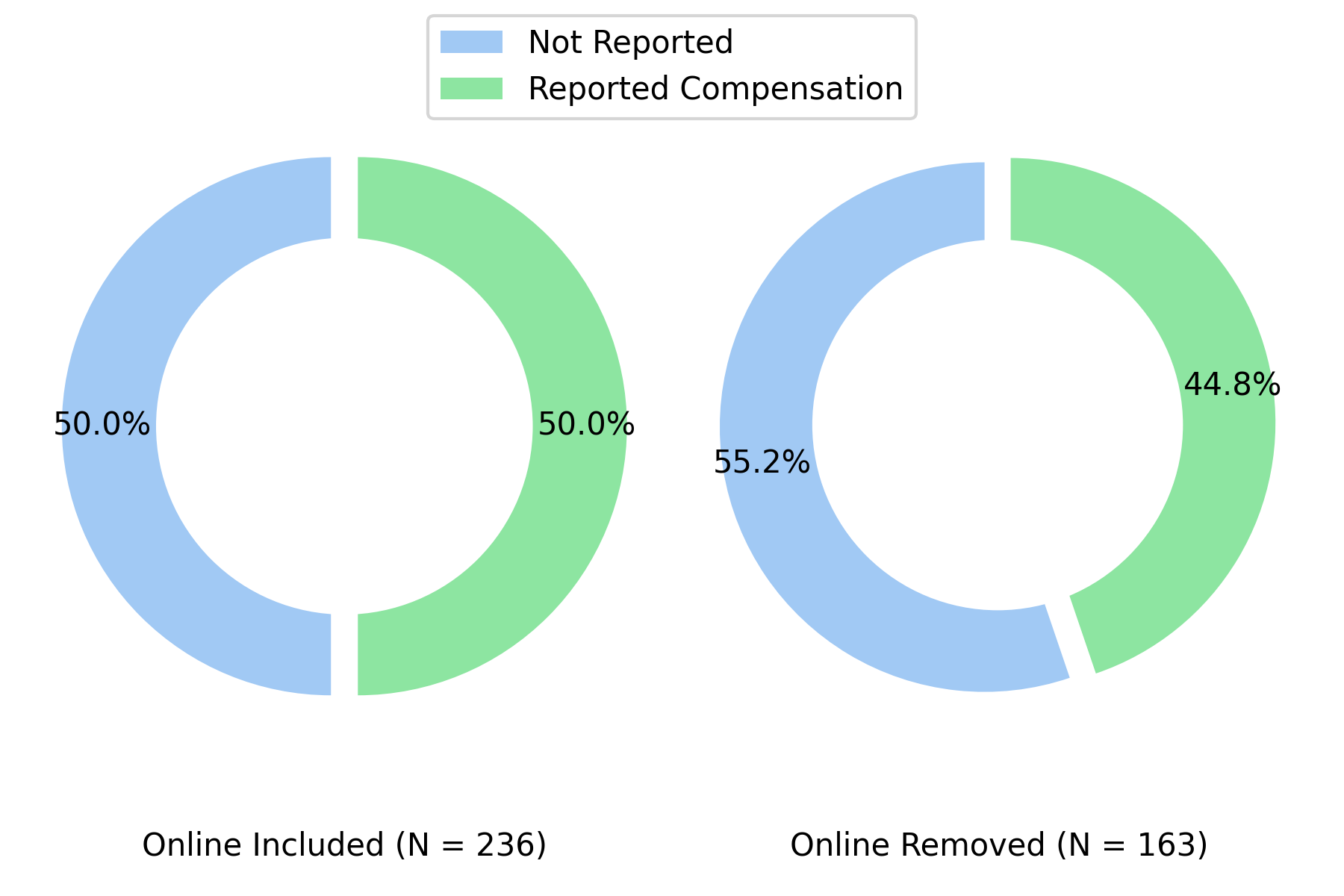}
  \caption{Papers that explicitly stated any direct benefit to the participants or a statement of no direct benefit and papers that did not.} 
  \label{fig:compensation} 
\end{figure}

 A total of 118 studies reported compensation. When online studies were removed, 73 studies reported compensation. Compensation reporting summaries can be found in Fig. \ref{fig:compensation}.

\begin{figure}[h] 
  \centering
  \includegraphics[width=\columnwidth]{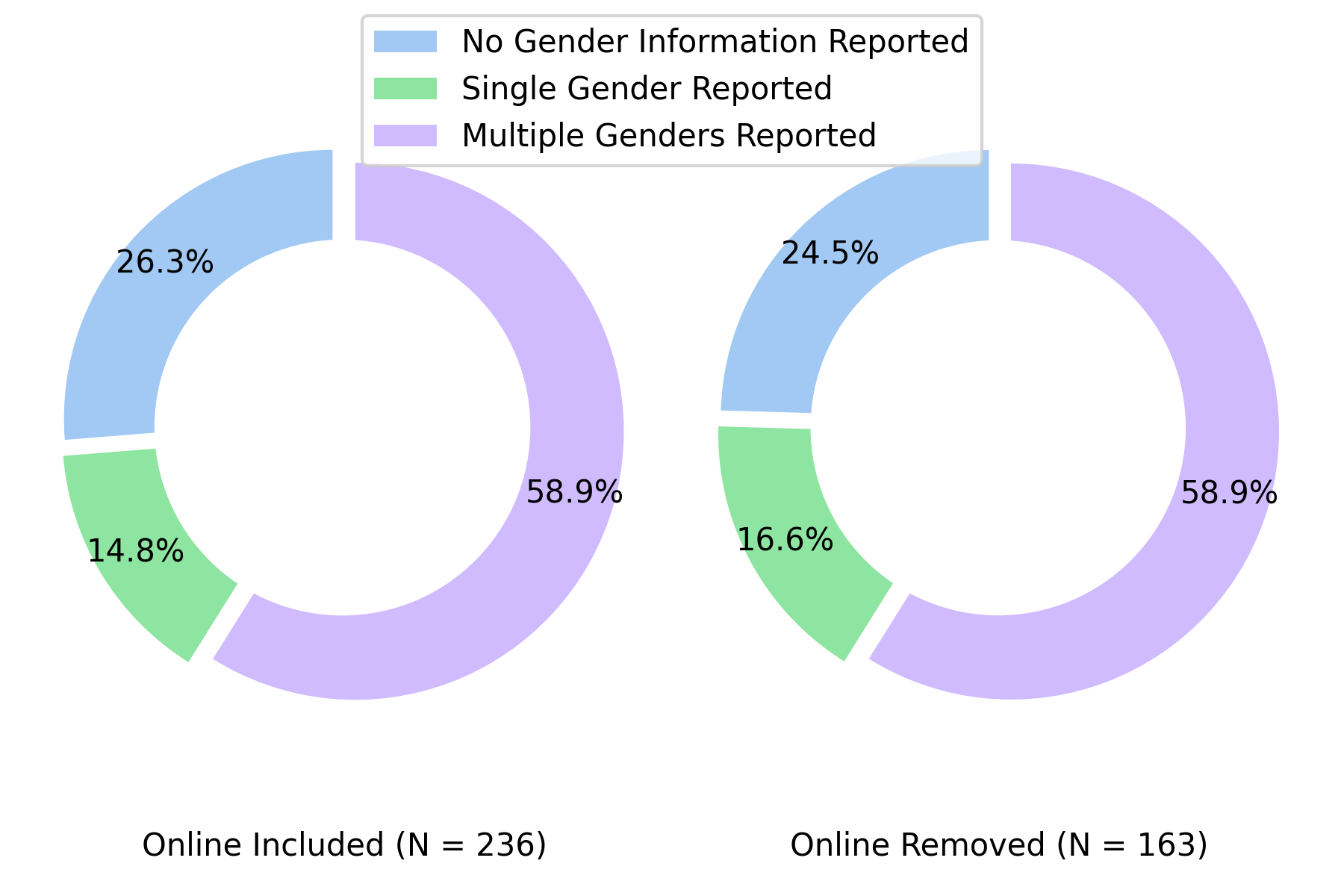}
  \caption{Papers that explicitly stated participant gender information and papers that did not. Papers that reported only a single gender (e.g., ``N=10; 5 participants were male'') are shown separately.} 
  \label{fig:gender} 
\end{figure}

 A total of 139 studies reported at least two or more genders, 35 only reported a single gender, and 62 did not report participant gender metadata. When online studies were removed, 96 studies reported two or more genders, 27 reported a single gender, and 40 did not report participant gender metadata. Gender reporting summaries found in Fig. \ref{fig:gender}.

\subsection{Between Conference Analysis}
\begin{table}[h]
\begin{center}
\begin{tabular}{l|llll}
Metadata Category            & $\chi^2$ & $p$            & Significance & DoF \\ \hline
Recruiment Method            & 11.95    & .003           & **           & 2   \\
Recruiment Method (N-OO)     & 0.42     & .810           &              & 2   \\
Ethics Board Approval        & 15.10    & \textless .001 & ***          & 2   \\
Ethics Board Approval (N-OO) & 9.62    & 0.008          & **           & 2   \\
Sampled Population           & 5.89     & .436           &              & 6   \\
Sampled Population (N-OO)    & 0.84     & .991           &              & 6   \\
Compensation                 & 0.88     & .642           &              & 2   \\
Compensation (N-OO)          & 2.10     & .349           &              & 2   \\
Gender                       & 6.62     & .158           &              & 4   \\
Gender (N-OO)                & 12.78    & .012           & *           & 4  
\end{tabular}
\caption{Chi-square test for independence \cite{mchugh2013chi} results between each conference (HRI'19, HRI'20, and HRI'21) for each category. N-OO indicates ``non-online only'' studies included within the test. Significance values used were $p < .05$ = *, $p < .01$ = **, $p < .001$ = ***.}
\label{tab:chisquare}
\end{center}
\end{table}

A summary of Chi-square tests for independence \cite{mchugh2013chi} among the conferences (HRI'19, HRI'20, and HRI'21) for each metadata category can be found in Table \ref{tab:chisquare}. The $p$ value measures the likelihood that the observed association between each independent variable (i.e., conference year) and the dependant variable (i.e., metadata category) is caused by chance.

A significant result was found for recruitment method with all studies ($p=.003$) but not when online studies were removed ($p^{N-OO} = .810$). A possible contributing factor was the imbalance of online to offline studies for each conference; HRI'21 contained a relatively high ratio of reported to not reported when online was included (42:31) and a much lower ratio when online studies were removed (10:24). HRI'19 (19:41 $\xrightarrow{}$ 12:37) and HRI'20 (36:67 $\xrightarrow{}$ 19:61) remained relatively consistent.

A significant result was found for ethics board approval for both sets of data ($p < .001, p^{N-OO} = .008$). HRI'19 had lower reported to not reported rate (16:44) as opposed to HRI'20 (57:46) and HRI'21 (41:32).

A significant result was found for gender reporting when removing online studies ($p^{N-OO} = .012$) but not for the all studies data set ($p = .157$). A possible contributing factor was the HRI'21 change of reporting when dropping online studies (40:12:21) $\xrightarrow{}$ (13:11:10) mapped as (Multiple Genders Reported : Single Gender Reported : No Gender Reported). HRI'19 (30:13:17) $\xrightarrow{}$ (27:8:14) and HRI'20 (69:10:24) $\xrightarrow{}$ (56:8:16) were comparably consistent when dropping online studies.


\section{Discussion and Limitations}
The focus of this paper is on the recommendations for reporting HRI study metadata (Sec. \ref{sec:reviewandrec}) and not to disparage the field through the data reported. The data are reported only to help support the idea that we, as HRI researchers, need to better report study metadata for more reproducible and generalizable studies. The goal is to help encourage replicability studies such as ``A Three-Site Reproduction of the Joint Simon Effect with the NAO Robo'' \cite{strait2020three} from HRI'20. With the HRI conference including a reproducbility track in 2020 \cite{hri2020mainproceedings}, we hope to see similar focus in other conferences and journals.

By no means are the recommendations a strict guideline to reporting metadata and are apt to change. Reviewing studies from the prior three years highlights the amount of nuance needed for different study categorizations and metadata. The changing landscape of user studies, such as the reproducibility crisis \cite{baker2016reproducibility} or the international SARS-CoV-2 pandemic, indicate a need to reevaluate recommendations year-over-year.  

Finally, while we have attempted to make a case for reporting guidelines in HRI, it is also important to acknowledge the limitations of such a review. We only sourced papers from the International Conference on Human-Robot Interaction (2019 through 2021); differing trends from the ones we report on may be discovered in a review of a different conferences or in journals. Additionally, as our review only covers the three most recent years, it is difficult to discern established trends in reporting HRI studies. Finally, we believe our coding is consistent and minimizes human error, but there is always possibility of human error in any manual search process.

\section{Conclusion}
Human participants are by definition a key component of human-robot interaction studies, and their behavior in studies will be at least in part impacted by the events they experience and their personal identities. Thus, to better support the reproducibility of studies and generalizability of a study’s results, it is critical that who participants are and what they experience throughout the course of a study are reported. In this paper, we reviewed the reporting guidelines for participant recruitment, compensation, and gender in the related fields of psychology, medicine, and HCI. We then surveyed the papers published in HRI 2019-2021 to examine reporting the same three categories and found that rates of reporting both individually important for understanding full study context, to be insufficient. We aim not to undermine the quality of any paper, but to suggest a higher rigor of reporting metadata in each study, including the authors’ own, to allow for greater reproduction of results in HRI.

\section*{Acknowledgment}
This work was supported by the National Science Foundation (NSF) under award IIS-1925083.

\bibliographystyle{ieeetr}
\bibliography{bibliograph}
\end{document}